# Visual Machine Learning: Insight through Eigenvectors, Chladni patterns and community detection in 2D particulate structures

Raj Kishore[1,2], S. Swayamjyoti[1,3], Shreeja Das[1], Ajay K. Gogineni[2], Zohar Nussinov[4], D. Solenov[5], Kisor K. Sahu[1,2*]

1. School of Minerals, Metallurgical and Materials Engineering, Indian Institute of Technology, Bhubaneswar-751007, India
2. Virtual and Augmented Reality Centre of Excellence, Indian Institute of Technology, Bhubaneswar-751007, India
3. NetTantra Technologies (India) Pvt. Ltd., Bhubaneswar, Odisha 751021, India
4. Department of Physics, Washington University in Saint Louis, MO- 63130-4899, USA
5. Department of Physics, St. Louis University, St. Louis, Missouri 63103, USA

*Correspondence to *kisorsahu@iitbbs.ac.in*

*Abstract*

Machine learning (ML) is quickly emerging as a powerful tool with diverse applications across an extremely broad spectrum of disciplines and commercial endeavors. Typically, ML is used as a black box that provides little illuminating rationalization of its output. In the current work, we aim to better understand the generic intuition underlying unsupervised ML with a focus on physical systems. The systems that are studied here as test cases comprise of six different 2-dimensional (2-D) particulate systems of different complexities. It is noted that the findings of this study are generic to any unsupervised ML problem and are not restricted to materials systems alone. Three rudimentary unsupervised ML techniques are employed on the adjacency (connectivity) matrix of the six studied systems: (i) using principal eigenvalue and eigenvectors of the adjacency matrix, (ii) spectral decomposition, and (iii) a Potts model based community detection technique in which a "modularity" function is maximized.  We demonstrate that, while solving a completely classical problem, ML technique produces features that are distinctly connected to quantum mechanical solutions. Dissecting these features help us to understand the deep connection between the classical non-linear world and the quantum mechanical linear world through the kaleidoscope of ML technique, which might have far reaching consequences both in the arena of physical sciences and ML.

## 1 Introduction

Advanced machine learning (ML) techniques have been developing at an unimaginable pace. It is primarily because of the success of ML techniques in many diverse areas of applications [1-6]. Today, there is hardly any scientific discipline that has not been influenced and improved by ML in one way or the other. However, the main concern is the lack of an in-depth understanding of how the desired results are being produced. For example, Neural Networks (NN), a form of supervised ML technique, were initially considered as black boxes since their functioning was not directly interpretable. However, it is now appreciated that NN can be considered as function approximators, where the network learns an internal representation or a function that maps from input to the output. In the case of image recognition, they perform better than traditional feature extraction techniques where each feature is computed based on specific domain knowledge [7, 8]. For example, a far more advanced variation of NN, Convolutional Neural Network (CNN), which was inspired by the image processing by the visual cortex, contains many hidden layers, where each hidden layer involves matrix multiplication with filters. A Filter is a square matrix of typical size 3 x 3 (say). The dot product is taken between the filter and all the possible 3 x 3 regions in the given images. The center pixels of the 3 x 3 regions in the image are replaced by the dot product. The numbers in these filters are the weights of CNN and are optimized through Back-propagation to meet the objective. There were several attempts to understand how the weights are changing as we go from one hidden layer to the next inside a CNN. It is done by visualizing the 'filters' and 'activations' being learned by the network [9, 10]. These visualizations help us in understanding the structure of the features learned by CNN. When the particular feature is present in the input image, a high activation is generated at the output. The layers hierarchically learn features. Initial layers are good at detecting edges in an image. As we go deeper into the network, complex features are learned on top of the low-level features learned initially. 'Saliency Maps' and 'Class Activation Maps' (CAM) computed by using final hidden layers of CNN help in knowing which region in the input image was important in predicting a certain output [11, 12]. Suppose that the NN predicted that an image contains a 'Dog', computing saliency maps can help us to get the location of 'Dog' in the image as described in [13]. However, a simple and comprehensive picture of "what's going on?" in an unsupervised ML study is still largely missing and deserves more clarity.

Thus the development of scientific intuition is of utmost importance to further accelerate the pace of innovation in this arena. The present article attempts to elucidate the inherent functioning of unsupervised learning methods, particularly the "community detection" technique. Moreover, in Sect. 4, it will be argued that the underlying mathematical fabric of both supervised and unsupervised ML techniques are not very different (at least from the viewpoint of developing basic scientific intuition). In most of the community detection problems, the first step is the abstraction of the problem statement in the form of a graph. The second step is to find the optimal partition of the graph [14-17]. This second step is the core of the unsupervised machine learning problems. The first step is generally problem-specific, and usually, a generic prescription is not available. Even then, some common features do exist that characterize the problem at the first stage itself, and three such features are the number of nodes (*N*), number of edges (*E*), and weight of the edges (*W*). Since the second step is an NP-hard problem [18, 19], an exact solution might be obtained by standard numerical methods of linear algebra, only when *N* is small. When *N* is very large, in general, the exact solution cannot be found in finite time [20]. In this particular regime, machine learning actually thrives by proposing a reasonable solution with finite amount of resources and computation time [21-24]. Before going into the complex and somewhat opaque community detection scheme, we will discuss and develop a systematic understanding of two exact models, which might be capable of solving networks of small sizes under favorable conditions. To this end, we will employ an exact method based on the calculation of eigenvalues and corresponding eigenvectors of the connection (or, adjacency) matrix of a given network [25, 26]. We choose this method, not because it is practically useful for large *N* (or *E*) problems, but because it can help to develop scientific intuition via visual representations. We will also discuss a spectral decomposition method [27-31] and will, finally, employ a community detection technique [17]. We will conclude with remarks on how these insights might be used to better understand ML algorithms (including aspects of deep learning networks).

It is interesting to note that the results of the exact diagonalization method, as we will see later in the results (Sect. 3.1.3 and 3.1.4), bear some similarities with quantum mechanical orbitals. While the problem in question is completely classical, features inherent to quantum mechanics emerge as part of the constructed description. Indeed, it will be demonstrated that the establishment of the underlying connections between the supervised, unsupervised and deep neural networks goes way beyond its initial scope and is connected to the more fundamental mysteries of nature. Classically the world around us (particularly the dynamics of many body systems) is non-linear. It is therefore not a surprise that the equation of motions that were developed from classical considerations are also non-linear. It took mankind quite some time to understand that, actually the deep underlying equations of motion, in particular, the Schrödinger equation, which is the quantum mechanical (QM) equation of motion and arguably the most fundamental equation of nature that describes any and every system around us is essentially linear. For large systems ( $\lim_{N \to \infty}$ ) most quantum effects disappear, though some quantum-like features can still be useful in understanding the problem as it occurs, e.g., in (classical) statistical mechanic invoking the notion of a classical state. To this end, a comprehensive understanding of quantum-like features in the solution of a ML problem can produce much illuminating insight.

## 2 Methods

In this work, we use different material systems as a case study. Yet, it is important to note that the insight gained from these systems is not restricted to materials systems only and can be generalized for other systems as well. Atoms in a material interact with the neighboring atoms. During the conversion of an atomic ensemble into a graph, the atoms are denoted by nodes and the interactions are represented by edges [32, 33]. Similarly, if we take a granular system, the centre of mass of each particle can be represented by a node and interactions between them by edges [34]. The major difference between these two systems is in the details of their interaction schemes: while in atomic system interactions can be attractive or repulsive; in case of granular systems, the primary mode of interaction is only repulsive [35, 36]. Since in this article, we have considered granular systems, an edge will represent physical contact between particles. As the machine learning schemes do not need details of contact mechanism (attractive vs. repulsive interaction), therefore for the sake of generality, we will not specify the exact mode of interaction for the machine learning purpose (yet, one can still explicitly differentiate between attractive and repulsive interaction in discussion).

We have studied six systems of different sizes and configurations shown in Fig. 1(a) to (f) respectively. In particular, Fig. 1(a) defines an extremely small and simple toy model; Fig. 1(b) - a little larger 2D packing; Fig. 1(c) - a 2-dimensional (2D) crystalline packing with no defects enclosed in a circular region; Fig. 1(d) - another defect-free 2D crystalline packing enclosed in a square region; Fig 1(e) - a 2D crystalline granular ensemble with multiple point defects, and Fig. 1(f) - a 2D crystalline granular packing having both point and line defects. It is to be noted that we are using 2D systems for this study so that we can easily depict and analyze the results and invoke intuition. However, all the methods discussed here are easily extendable to 3D systems also (or n-D in general). In the following, we discuss and construct the protocols.

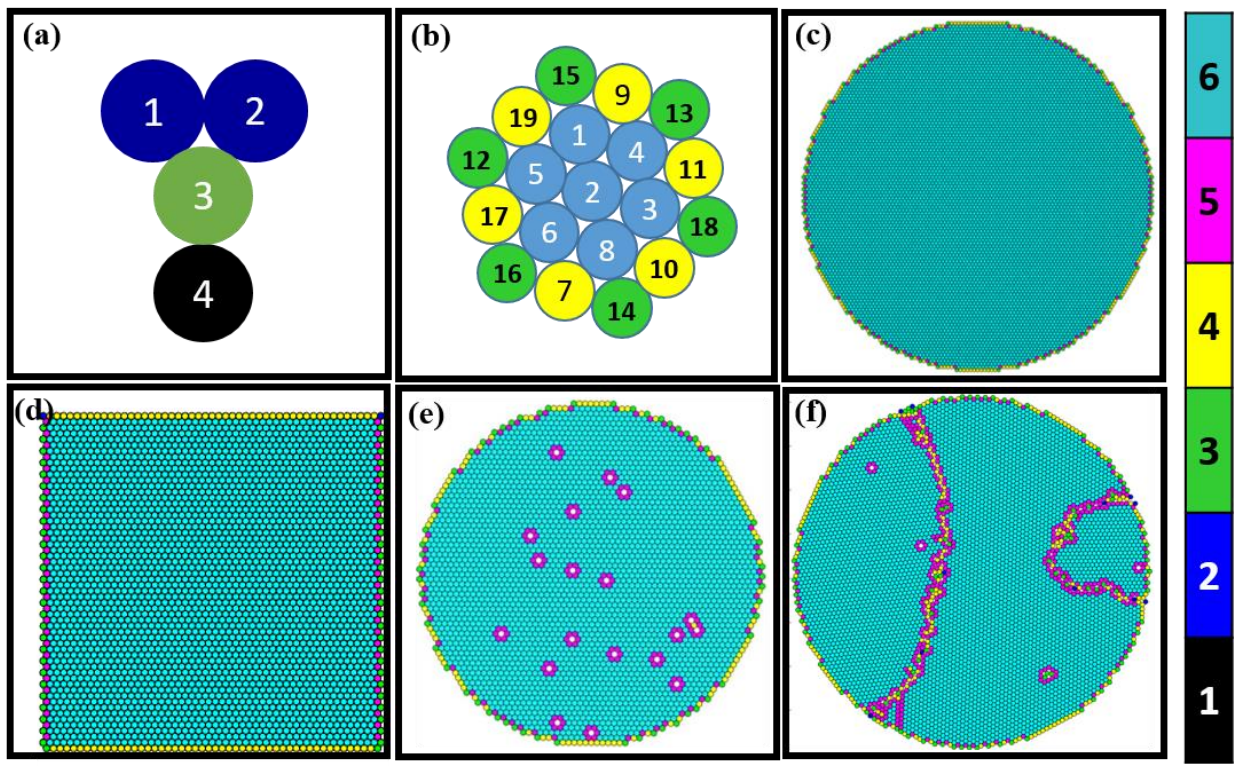

**Fig. 1** Different network systems studied in this article. (a) An extremely small and simple hand-made toy network (b) a small 2D packing and (c-f) are different granular assemblies obtained by physics-inspired DEM simulation of centripetal packing. Details of the simulation protocols are available in supplementary materials. Figures (a) to (f) will be denoted as Structures I to VI, respectively, throughout this article. The color bar in extreme right represents coordination number for panel (a) to (f)

Figure 1(a) shows a hand-crafted small toy network, which will be referred to as Structure-I. This model is so small that it is irrelevant to ask whether the model is 2D or 3D. Since the machine learning models used in this article depends only on the abstracted network, the true underlying Euclidian dimensionality of the problem will not make any difference in the results (because of topological invariance). Figure 1(b), or Structure-II, depicts a somewhat larger (than Structure-I) 2D hand-crafted packing. Although Structure-II apparently resembles a 2D crystalline packing (hexagonal), in reality, it has more atoms in defect positions (12 atoms on the boundary) than in crystalline positions (7 atoms in hexagonal interior 'lattice'-like positions). It is to be noted that if we assume the atoms in Structure-II to be cohesive (having attractive force field), then transitioning this 2D structure to 3D will modify the connectivity (adjacency) matrix, and therefore, the results will be different. Therefore, this structure can be thought of as a 'critical' size structure, where dimensionality makes a difference to the result (assuming that a change in dimensionality leads to change in connectivity and, thus, topology). Incidentally,

Structures III, V, and VI, corresponding to Fig. 1(c), (e) and (f) are generated using a physics-guided granular packing protocol called Discrete Element Modeling (DEM) [37, 38]. It is described in detail in the supplementary materials (S1). Structure-IV, see Fig. 1(d), is prepared using a basic crystallographic protocol to generate regular hexagonal packing. It is to be noted that, one could have easily made packing similar to Fig. 1(c), (e), and (f) using only crystalline packing protocol (without explicitly using any physics/chemistry rules).

Since both, the exact method based on eigenvector corresponding to the largest eigenvalue, and the community detection method, depend on the adjacency (connectivity) matrix of these structures, we depict them in Fig. 2 for all investigated structures. Figure 2(a) depicts the adjacency matrix both in numerical form (first panel) and in the encoded image form (second panel). The elements of the adjacency matrix, $A(i,j)$ are unity if nodes $i$ and $j$ are in contact or zero otherwise. In the image encoding of this adjacency matrix, white and black pixels represent 1 and 0 respectively. For Structures III-VI, the size of the adjacency matrix is very large (minimum size is 2800×2800), and only image encoding is presented, see Figs. 2(b) through (f). It may be noted here that there is no loss of information in this image encoding.

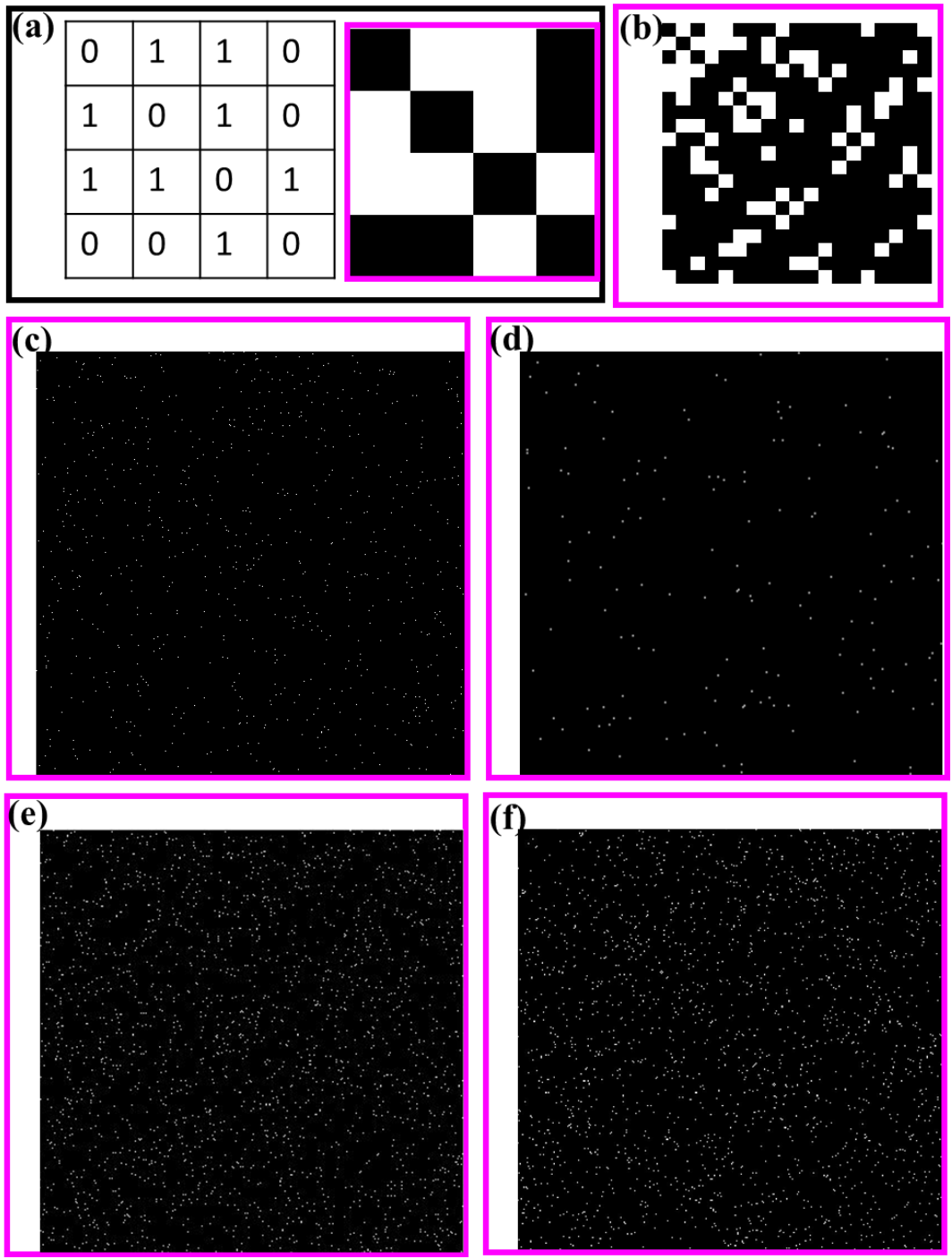

**Fig. 2** Adjacency matrix of all the six systems studied. The first panel of (a) explicitly shows the elements of the matrix, and the second panel of (a) shows its image representation, where 0 and 1 are denoted by black and white "pixels," respectively. Since for Structures III-VI, the adjacency matrices are large; they are represented only by the corresponding "image" representation through (b) to (f). The network obtained from structure III-VI are sparse networks, thus to show some white pixels, we have shown only the first quadrant of the image.

# 3 Results

The results in this article are presented in three sub-sections. Part-A contains the results for structures I-VI using the exact technique. Part-B depicts the results using the spectral decomposition method. Part-C contains the results based on community detection.

## 3.1 Part-A: Results based on the exact method

In this section, we discuss how machine learning results can be obtained by using exact methods. Though this type of approach is quite restrictive and is not particularly favorable for machine learning operations involving large systems, they are useful for developing an understanding of the results. Adjacency matrix $A$ has an associated eigenvalue equation of the form: $AX=\lambda X$, where $\lambda$ is eigenvalue, and $X$ is the corresponding right eigenvector. Throughout this article, we are explicitly dealing with the right eigenvectors. By construction, the adjacency matrix ($A$) is square and symmetric, and therefore, the right and left eigenvectors are identical. A very important distinction from the main-stream supervised learning schemes needs to be made at this point. Most generically, the data for supervised learning (both training and test data) are not square, let alone being symmetric. Additionally, negative entries are also not restricted. So the right eigenvectors and left eigenvectors are typically not identical for those supervised learning cases. However, the majority of unsupervised learning problems are typically concerned with "clustering" or "community detection" of the input data set. For those cases, the first step is to construct a network, whose adjacency matrix, by construction, is square and symmetric (provided the network is non-directed). Additionally, $A$ in most cases have *all* non-negative entries; though one must appreciate that, this is not a strict requirement. So is the case for the present article. Henceforth, in this article we will call the right eigenvectors as simple eigenvectors.

The following is the plan for developing a step-by-step understanding of the mathematical states with increasing levels of complexity of the unsupervised machine learning method:

**Table 1** Structures arranged with increasing complexity levels and objectives of the detailed analysis for them

| Systems | Objectives |
| --- | --- |
| Structure-I | Notion of node-centrality (importance of a node in a network) |
| Structure-II | Classification based on 'Node-Centrality' (NC) |
| Structure-III | Further understanding of NC in bigger systems and importance of other eigenvectors |
| Structure-IV | Connection between the solutions of wave equations and eigenvectors |
| Structure-V | Presence of small (point) defects and its influence on NC |
| Structure-VI | Effect of larger (line) defects and insight for complex networks |

### 3.1.1 Eigenvectors of Structure-I: Notion of node-centrality (importance of a node in a network)

Since the size of this system is very small (*N*=4 nodes), we can easily depict all the eigenvalues (shown in Fig. 3(a) and all the corresponding eigenvectors. It is to be noted here that, while the adjacency matrix has no negative entries (they are either 0 or 1), still this is neither a positive definite nor a positive semi-definite matrix since it has negative eigenvalues. As illustrated in Fig. 3(a), there will be many eigenvalues, and the count will depend on system size (*N*). Of particular interest are the eigenvalues that have corresponding non-zero eigenvectors. All the eigenvectors for Structure-I are depicted in Fig. 3(b) and the eigenvector corresponding to the largest eigenvalue (which is 2.17) is highlighted by the green box. This eigenvector is also called the principal eigenvector. The entries of the principal eigenvector are mapped onto respective nodes and depicted in Fig. 3(c). It will be argued later that this is a measure of node centrality [39], and it can be exploited in machine learning (suitable for small systems).

In a mechanical rendition, the adjacency matrix may represent a network of springs. In such a mechanical analogue, a matrix element $A_{ij}=1$ may represent a spring connecting sites *i* and *j*. In such a case, the eigenvalues of A will determine the evolution of the spring network. Formally stated, the adjacency matrix *A* of such a system may be viewed as scaled "Hessian" or "dynamical matrix" whose eigenmodes correspond to decoupled mechanical vibration modes and eigenvalues determine the frequencies of these eigenmodes. In the limit of large system size, a network with springs between neighboring sites may represent an elastic medium and the corresponding eigenvectors determine the elastic (phonon/sound) modes. Since *A* is a real semi-positive square symmetric matrix the Perron–Frobenius theorem [40] mandates that *A* must possess a unique largest eigenvalue, also called principal eigenvalue, and the corresponding eigenvector will have all non-negative entries. Therefore the components of this eigenvector can be used to formulate different centrality measures, where centrality refers to the importance of the node in a network. Since node 3 in Fig. 3(c), has the highest node centrality value (0.612), it can be referred to as the most central node in this network. It can be noted here that Katz centrality [41] or Google's © page-rank [42, 43] for a website are also different forms of node centrality. It is to be further noted here that "node centrality" is not a unique notion and can be constructed upon various properties of the network, and therefore, the same node in a network can have different values/rank in those "centrality" measures. For example, a network can denote flow/traffic (for example, airline connectivity network) or contrastingly it can stand to denote cohesiveness between atoms in a material. The two ideas will lead to different centrality measures and values.

From the machine learning perspective, estimating an exact value of node-centrality is typically not the goal. Because it is likely to overfit and typically not good for generalization. Most important idea is to obtain a relative difference in the centrality measures. It can be easily appreciated that node-centrality values for nodes 1, 2 and 3 are close, while the value for node 4 is far off. We will return to this point when discussing the results for community detection in Sect. 3.3.

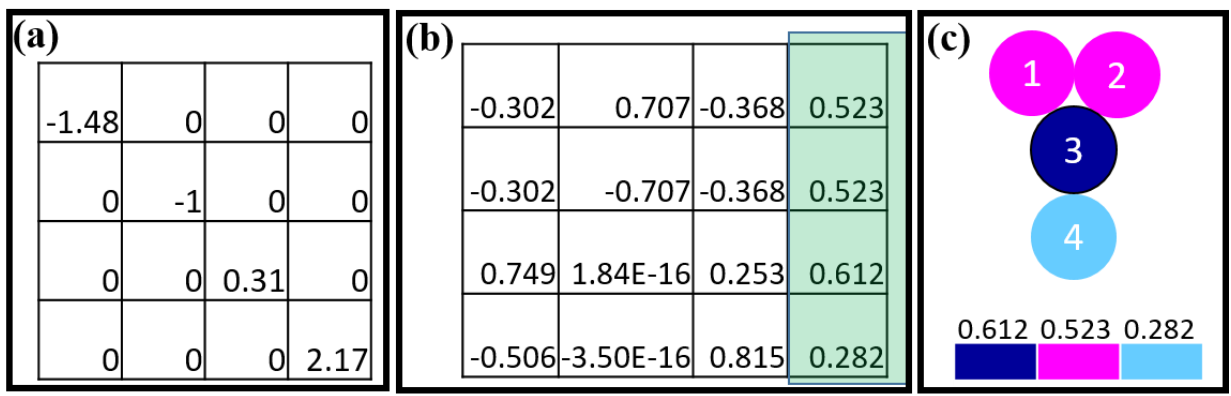

**Fig. 3** Pictorial representation of eigenvalues and eigenvectors of the adjacency (connection) matrix of a network. (a) The eigenvalues of the given connection matrix in Fig. 2(a) and (b) the corresponding eigenvectors. The principal eigenvector is highlighted with green color. (c) The given network, shown with the individual elements of principal eigenvector

### 3.1.2 Principal eigenvector of Structure-II: Classification based on 'Node-Centrality' (NC)

The principal eigenvector for Structure-II is: [0.30, 0.37, 0.30, 0.30, 0.30, 0.30, 0.18, 0.30, …….]. It is graphically represented by a color-coded mapping of the components of principal eigenvector onto each node in Fig. 4(a). Node 2 is the central atom and has the highest value of 0.37 (the largest component). All the orange-colored nodes (i.e., node #1, 3,4,5,6 and 8) have value of 0.30 (second-largest component). Grey colored nodes (node #7, 9,10,11,17 and 19) have value of 0.18 (third largest). Magenta nodes (node # 8, 12,13,14,16 and 18) have the lowest value of 0.13. The corresponding frequency distribution is plotted in Fig. 4(b). The node numbering scheme is deliberately chosen arbitrarily (not in a particular order) to point out that this method is not really dependent on the numbering scheme. This is important because generically the clustering or community detection schemes (or even stochastic gradient descent scheme, a popular method in supervised learning scheme) are not completely immune from similar arbitrariness. Though in Sect. 4, it will be argued that such dependence does not necessarily lead to inferior results. The largest value for node 2, denotes its importance in the centrality measures, followed by orange nodes and so on and this scheme can be used for classification. It can be noted that, if Structure-II is really extended to infinitely on crystal lattices, then all the nodes will turn out to be equivalent and will have the same centrality measure, meaning that no proper classification exists, which is indeed the right result because of the true equivalence of all the 'lattice' sites (contrast to the 'lattice'-like position in Fig. 1(b)).

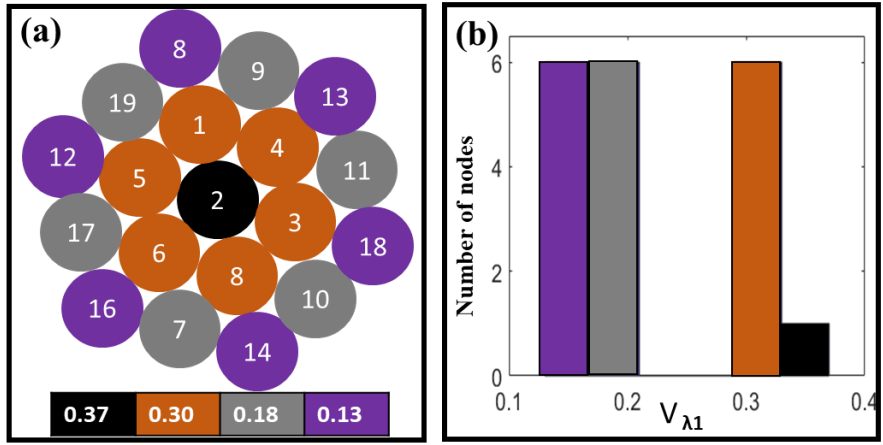

**Fig. 4** Pictorial representation of principal eigenvectors of Structure-II. (a) The assembly of particles with their particle ID written at the center. The particles are colored based on the individual elements of the principal eigenvector of the adjacency matrix (the color bar is given at the bottom) and; (b) the histogram showing the binning of elements of the principal eigenvector

### 3.1.3 Eigenvector of Structure-III: Further understanding of NC in bigger systems and the importance of other eigenvectors

The principal eigenvector for Structure-III is segregated into ten equal bins (chosen for convenience) and the frequency distributions are depicted in Fig. 5(b). The nodes are color-coded according to those bin values and are mapped onto the original structure and depicted in Fig. 5(a). Node-centralities are obvious by the radial pattern.

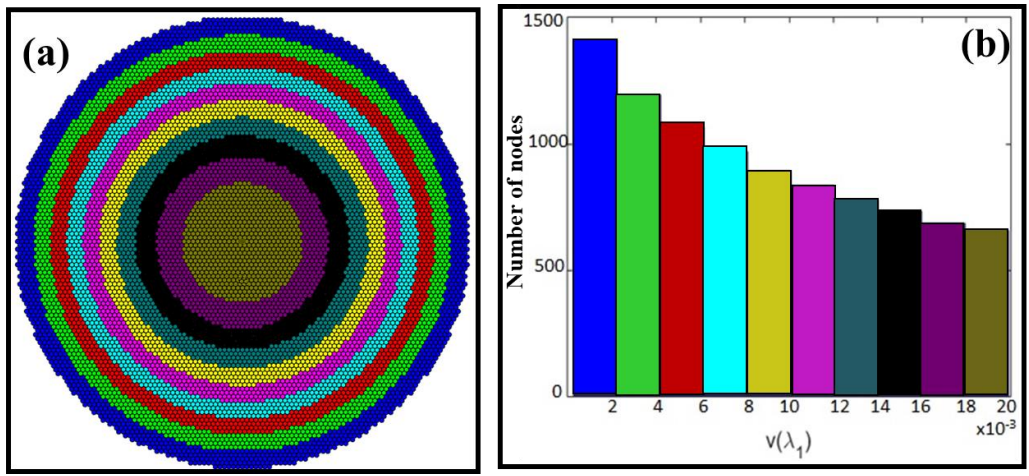

**Fig. 5** Visual analysis of principal eigenvector of the adjacency matrix obtained from a nearly 6-regular network (except for the boundary node, all the inner nodes have degree six). (a) The given granular assembly of 9200 particles, the particles are colored based on their corresponding elements of principal eigenvector. Color is chosen the same as the color of each histogram bins shown in panel (b). The frequency distribution of 9200 elements of principal eigenvector, which are divided into 10 equal bins

We now turn our attention to other eigenvectors. Though the principal eigenvector stands out for its clear physical interpretation, which is routinely taken advantage of in many machine learning applications, other eigenvectors also contain some important features which can shed additional light for a better in-depth understanding. It is to be noted that all the eigenvectors are mutually orthogonal. Therefore taking the inner (dot) product of any two eigenvectors will yield zero. If one, for some reason, needs to create a large number of orthogonal basis sets for the problem under study, these different eigenvectors will be the simplest choice. Figure 6 depicts all the eigenvectors corresponding to the 12 largest odd eigenvalues ($\lambda_1$ to $\lambda_{23}$) (the even eigenvectors have the same patterns as of odd eigenvectors, except for the orientation. The even eigenvectors are shown in supplementary S2). Reference values for color-coding of nodes in each panel of Fig. 6 are provided in the supplementary S3 section. One might argue that this choice could have been easily made using symmetry argument alone, without going through the eigenvector calculation. While that is true for very symmetric and regular cases (like the Structure-III), it will not be possible to guess such an easy solution if the structure has various defects present in them in different locations as exemplified by Structure-V and VI.

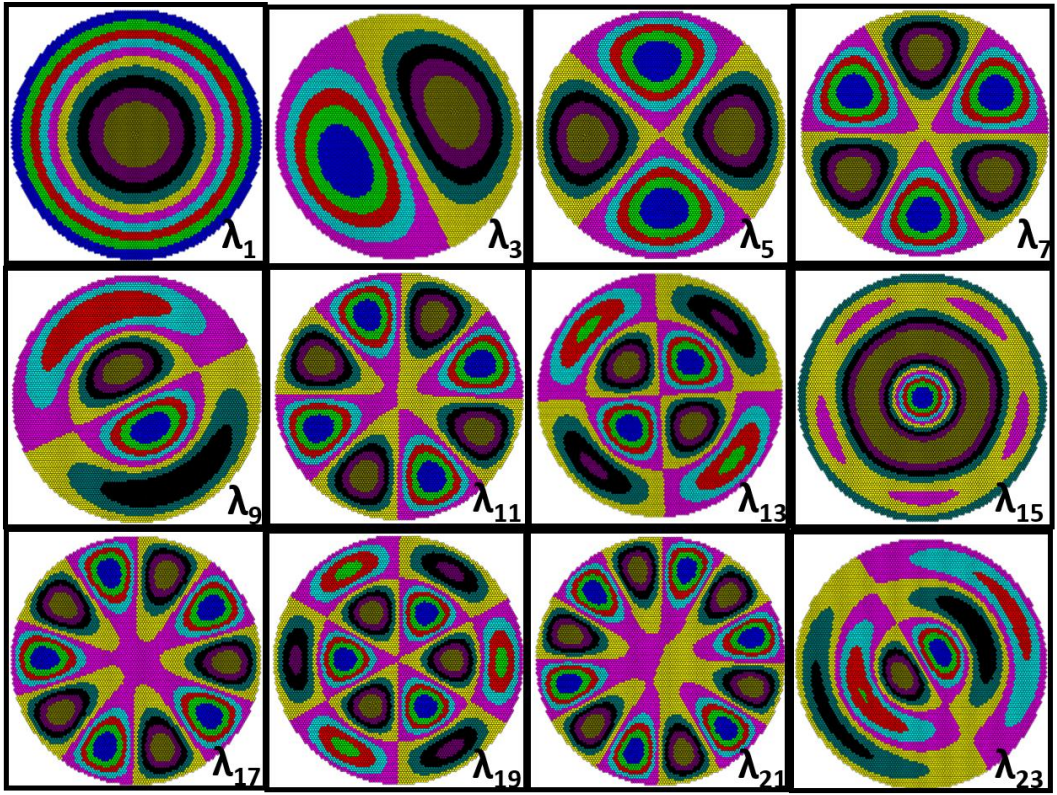

**Fig. 6** The visual representation of eigenvectors corresponding to other higher eigenvalues. The bin values of each eigenvector are shown in the table S3.1. All the eigenvectors are orthonormal to each other. Colormap (color-coding values) for this figure can be found in supplementary S3

### 3.1.4 Results for Structure-IV: Connection to solutions of the wave equation

In this section, we present the results for Structure-IV and establish its connection to the wave equation for identical structure (to that of a 2D plate). It should be noted that, had we considered a 3D structure, then the connection would have been with a 3D standing wave, and this will be discussed later. Equation (1) denotes a 2D wave equation on a 2D plate in the Cartesian coordinate system, chosen because of the geometrical consideration (square plate) [44].

$$\frac{1}{c^2}\frac{\partial^2 U(x,y,t)}{\partial t^2} = \frac{\partial^2 U(x,y,t)}{\partial x^2} + \frac{\partial^2 U(x,y,t)}{\partial y^2} \tag{1}$$

where, $U(x, y, t)$ denotes the local displacement in the normal direction (can be considered in the z-direction; though, strictly speaking, this is a 2D problem) at point $(x, y)$ at any given time $t$. This equation is useful for both the transient wave (changing with time) and well as for stationary waves (invariant with time). While the former can provide important insight into the dynamics of the system, it is the latter that is of particular interest to us. This is so because, using machine learning parleys, it provides us access to the ground truth that is not changing with time. Even in other areas of science, if a quantity is invariant with time, then most likely one can formulate a conservation law for it. Therefore, in the following, we discuss the stationary solutions for the wave equation and compare it with eigenvectors. We are particularly interested in the problem where, initially the displacement and velocity (the first derivative of displacement) are zero, as represented by Eqs. (2), and (3) and throughout the boundary, the displacement is also zero for all time, represented by Eq. (4).

$$U(x, y, 0) = 0, \quad 0 < x < a, \quad 0 < y < b \tag{2}$$

$$\frac{\partial U}{\partial t}(x, y, 0) = 0, \quad 0 < x < a, \quad 0 < y < b \tag{3}$$

$$U(\Delta\Omega, t) = 0, \quad t > 0 \tag{4}$$

For this problem, the solutions can be represented by Eq. (5)

$$C\sin\left(\frac{m\pi x}{a}\right)\sin\left(\frac{m\pi y}{b}\right) + D\sin\left(\frac{p\pi x}{a}\right)\sin\left(\frac{q\pi y}{b}\right) = 0 \tag{5}$$

where $\Delta\Omega$ represents the boundary, and we assume that the square plate has dimension $a=b=1$ (normalized).

In Fig. 7, the panels in the first column represent some selected eigenvectors for Structure-IV, while the panels in the second column represent the matching solutions of the wavefunction with suitable values for $m, n, p,$ and $q$. The similarities are expected because if we see the eigenvalue equation $AX=\lambda X$, then we should appreciate that the eigenvector are the special vectors that are invariant under the transformation by $A$. Therefore, the eigenvectors also represent the invariant modes much like the stationary state solutions. Incidentally, the plots in the second column of Fig. 7 are called Chaldni plots, named after Ernst Chladni (1756-1827), who was investigating different modes of sound waves in 2D plates [45-47]. It is worthwhile to mention here that when quantum mechanics was developed in the early part of the last century, the solutions of the wave equation (standing wave modes) were readily available, because they were nothing but 3D analogue of the Chladni plots. Indeed, electron orbitals are the representation of Chladni plots in 3D. It is therefore not a surprise that, had we studied 3D structures, we would have recovered electron-obital like eigenvalue structures.

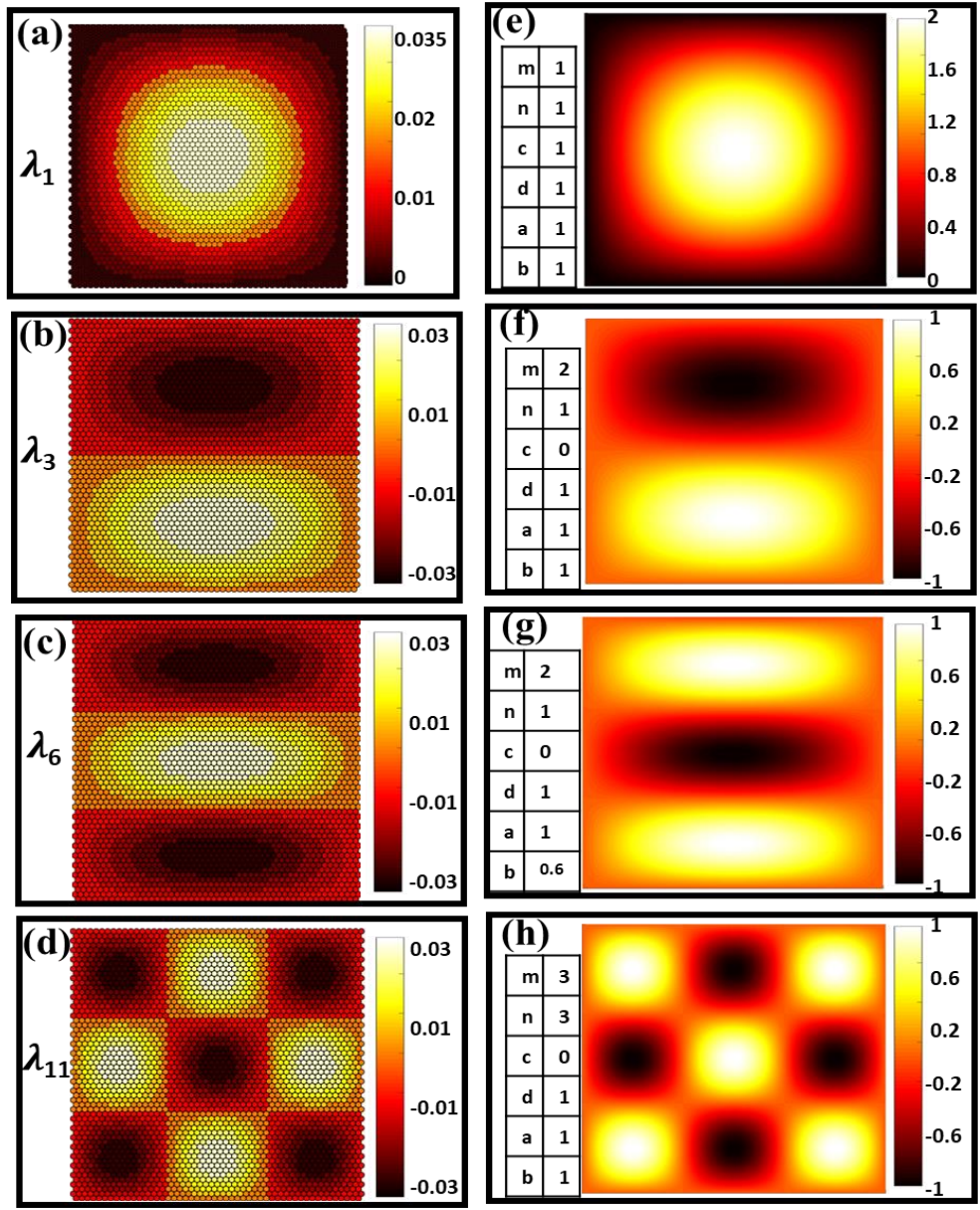

**Fig. 7** Comparison of different eigenvectors of Structure-IV with the different modes of the wave equation. The visual representation of eigenvectors for Structure-IV for eigenvalues (a) $\lambda_1$, (b) $\lambda_3$, (c) $\lambda_6$, (d) $\lambda_{11}$ ($\lambda_1$ is the principal eigenvalue and $\lambda_3$, $\lambda_6$, $\lambda_{11}$ are the 3rd, 6th and 11th highest eigenvalue respectively). (e-h) The matching solutions of the wavefunction (Eq.1) with suitable values for *m, n, p* and *q* (Eq. 5)

## 3.1.5 Principal eigenvector of Structure-V: Presence of small (point) defects and its influence on NC

We have studied the effect of point defects in the granular assemblies on the principal eigenvector to see how mode structures are distorting to accommodate the influence of the defects.

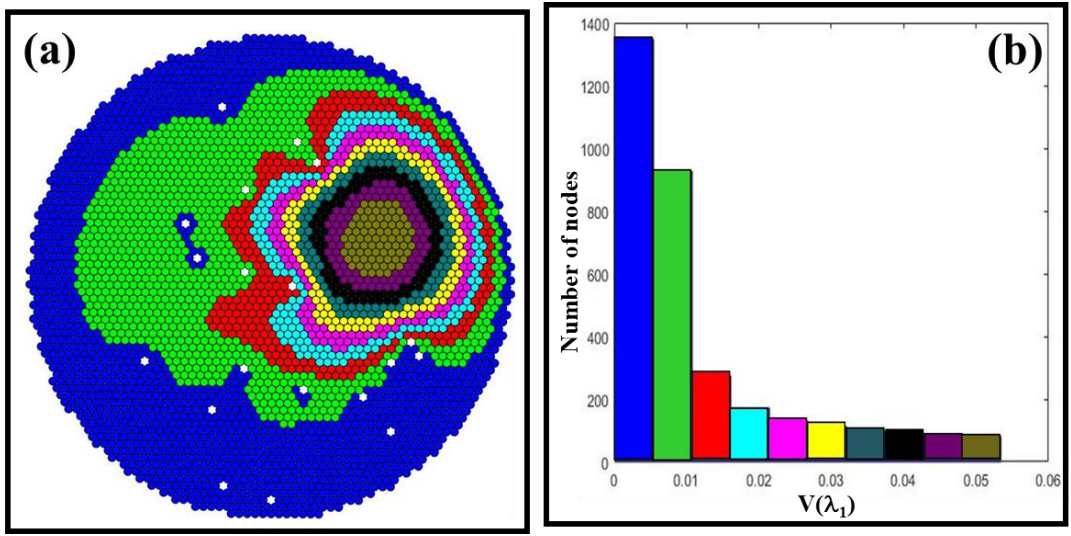

**Fig. 8** The visual analysis of principal eigenvector of the adjacency matrix obtained from a granular ensemble that has point defects/disorder. (a) The given granular assembly of 3350 particles, the particles are colored based on their corresponding principal eigenvector values. Color is chosen the same as the color of each histogram bins shown in panel (b). The 3350 elements of principal eigenvector are divided into 10 bins

## 3.1.6 Principal eigenvector of Structure-VI: Effect of larger (line) defects and insight for complex networks

Figure 9(a) shows the principal eigenvector for Structure-VI. As we see, most of the major modes are confined in the central part, which is clearly separated by strong line defects. Figure 9(b) represents the corresponding frequency distribution. It is, however, worth mentioning that side portions of the Structure-VI has different modes as shown in Fig. 9(c) and corresponding frequency distribution is shown in Fig. 9(d). One can make out that they are very weakly coupled with the central portion.
14

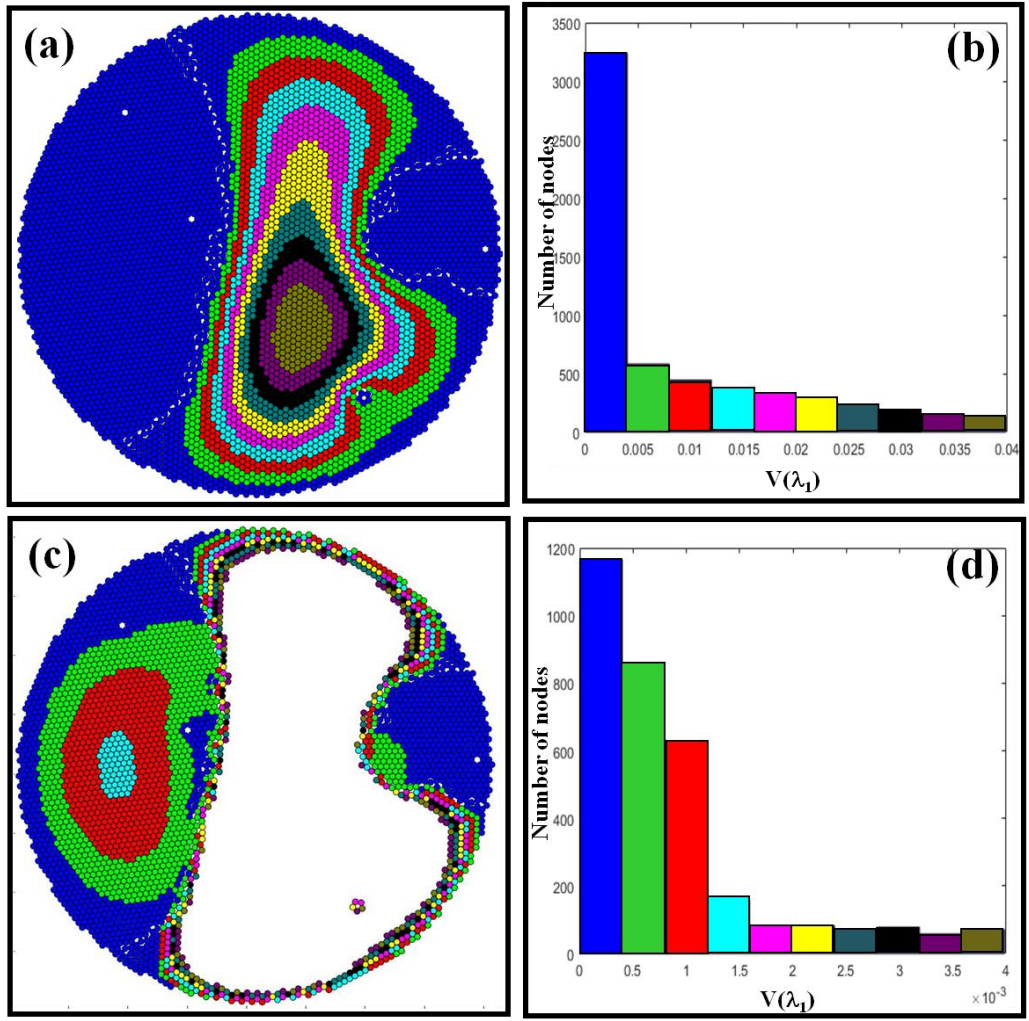

**Fig. 9** The visual analysis of principal eigenvector of the adjacency matrix obtained from a granular ensemble that has larger line defects/disorder (grain boundary like structure). (a) The given granular assembly of 5865 particles, the particles are colored based on their corresponding principal eigenvector values. Colors of particles are chosen to be the same as the color of each histogram bins shown in panel (b). The 5865 elements of principal eigenvector are distributed in 10 bins. (c) The side lobes also have a weak variation, which can be seen only if the central lobe is not considered. (d) The frequency distribution corresponding to (c).

### 3.2 Part-B: Some results by spectral methods

Spectral clustering is a technique of finding communities of nodes inside the network based on the node connectivity. It uses spectrum (eigenvalues) of the adjacency matrix of the network. We have used the community detection method using eigenvectors discussed in [48]. The steps of community detection are given below:

- Compute the eigenvalue and eigenvectors of the adjacency matrix of the network.

- Analyze the elements of the principal eigenvector, and if all the entries are positive, then it shows that all the nodes of the network belong to one community. If some entries are positive, and some are negative (note: 0 is considered as a positive entry), then the nodes should be divided into two groups based on their eigenvector values.

- We repeat the above interpretation to all the subsequent eigenvectors (in the decreasing order of the corresponding Eigenvalues) and then find out the smaller communities inside the larger components.

- Modularity of the components is used to decide whether it should be break into two smaller communities or not. If the modularity of the bigger component is less than the sum of the modularity of two smaller communities, then it should break into two communities else not.

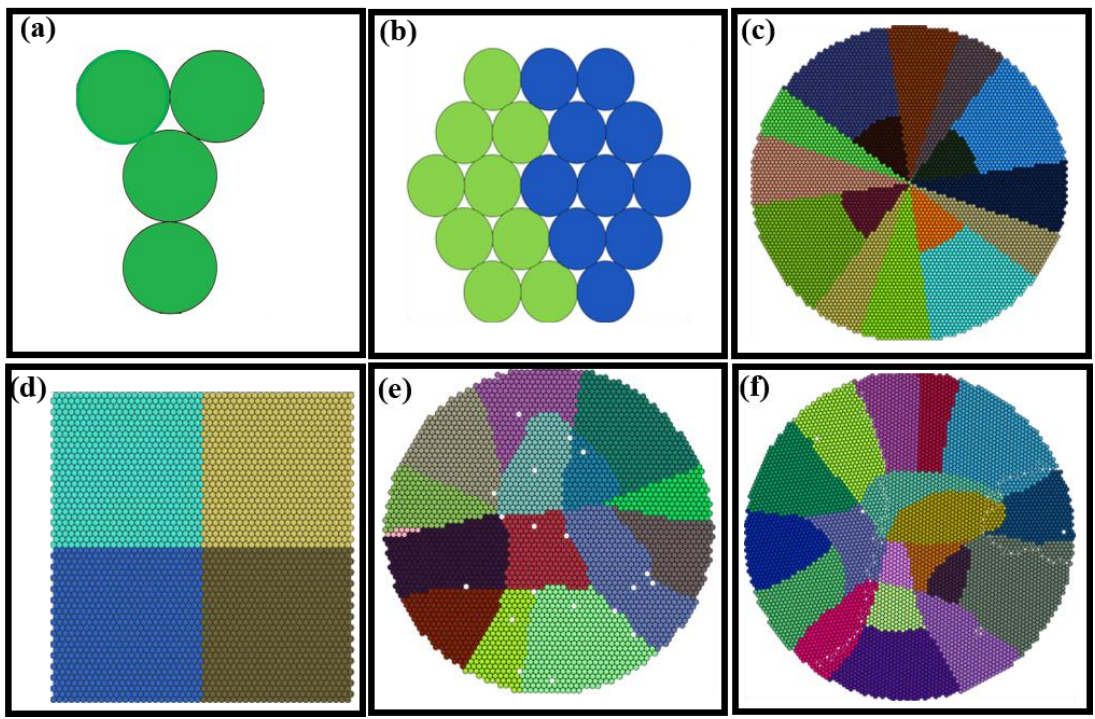

**Fig. 10** Communities detected in all the six structures (a-f) through spectral clustering method (color to different communities are randomly assigned)

Figure 10 depicts that the spectral clustering method produces sub-optimal results for all the six Structures (I to VI). Therefore it is not further elaborated here. We should note that there are more sophisticated (and accurate) spectral methods such as those in [49, 50].

### 3.3 Part-C: Some relevant results obtained by community detection

We have clustered all the six configurations (Fig.1) using the modularity maximization method. The clustered structures are given in Fig. (11). We have maximized the modularity function, which is developed for clustering the spatially embedded networks [17] and given in Eq. (6).

$$Q(\sigma) = \frac{1}{2m} \sum_{i,j} \left( a_{ij} A_{ij} - \theta(\Delta x_{ij}) |b_{ij}| J_{ij} \right) \left( 2\delta(\sigma_i, \sigma_j) - 1 \right) \tag{6}$$

Here $a_{ij}$ and $b_{ij}$ are the strength (not to be confused with edge weights) of connected and missing edges between the $i^{th}$ and $j^{th}$ nodes respectively. For setting up the strengths of edges, $a_{ij}$ and $b_{ij}$, there can be many choices. Since, in this study, we are dealing with unweighted networks, the strength $a_{ij}$ and $b_{ij}$ can be calculated as

$$a_{ij} = b_{ij} = \left( \frac{k_i + k_j}{2} - <k> \right) \tag{7}$$

The average degree of the network $<k>$ is given as

$$<k> = \frac{1}{N} \sum_{r=1}^{N} k_r \tag{8}$$

$k_r$ is the degree of $r^{th}$ node and $N$ is the total number of nodes in the graph. The function compares the local degree distribution at node level with the average degree distribution of the network as given in Eq. (7). Its value will be high if the nodes of a community are highly linked with each other. Highly linked communities exhibit a local degree distribution that is larger than the average degree distribution over the entire network.

$$\{2\delta(\sigma_i, \sigma_j) - 1\} = \begin{cases} 1 & \sigma_i = \sigma_j \ (i.e \text{ in same community}) \\ -1 & \sigma_i \neq \sigma_j \ (i.e. \text{ in different communities}) \end{cases} \tag{9}$$

We have penalized the intercommunity edges (Eq. 9) and appreciated for intra-community edges. Heaviside unit step function $\theta(\Delta x_{ij})$ which incorporates the geometrical constraints in the form of neighborhood definition. Here $\Delta x_{ij} = x_c - |\vec{r}_i - \vec{r}_j|$ is the difference in Euclidian distance between nodes $i, j$ and $x_c$ defines cutoff distance for neighborhood

$$\theta(\Delta x_{ij}) = \begin{cases} 1 & \Delta x_{ij} > 0 \ (\text{within cut off}) \\ 0 & \text{otherwise (outside cut off)} \end{cases} \tag{10}$$

The complete description of the given clustering method can be found in [17]

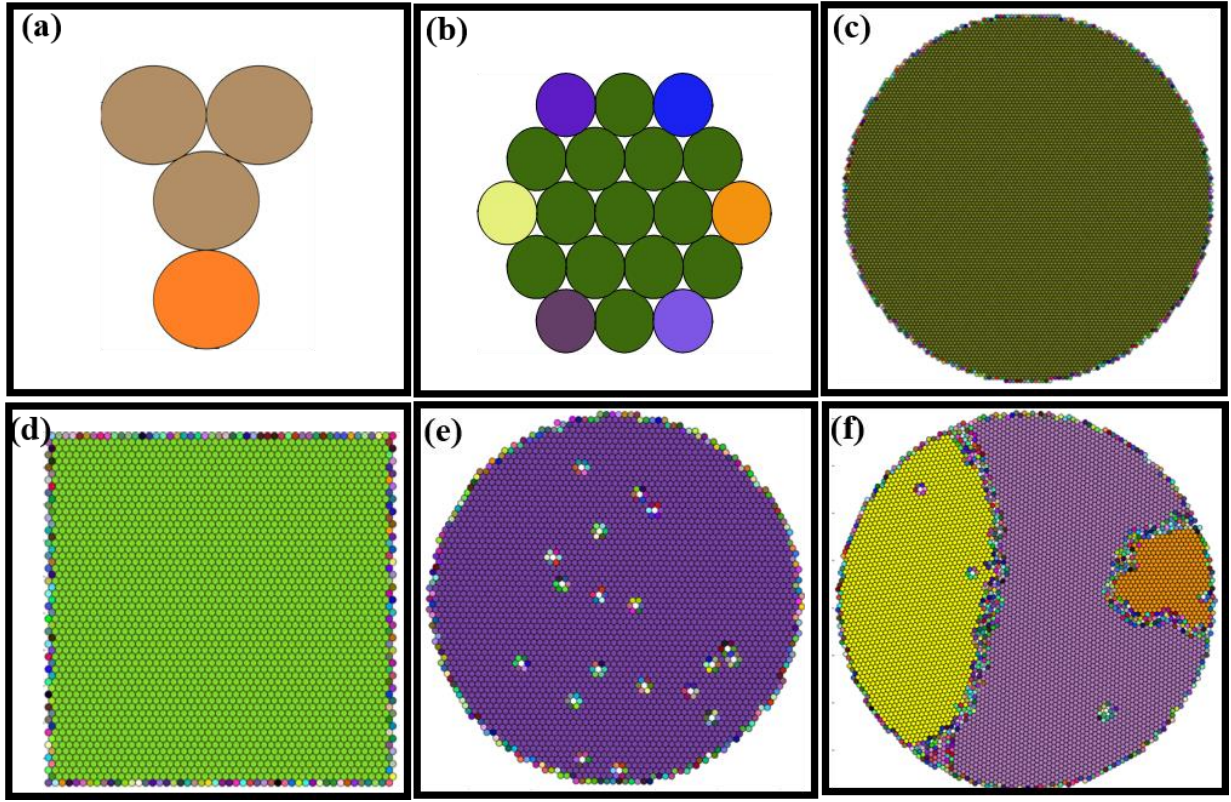

**Fig. 11** Communities detected in all the six structures (a-f) through maximization of modularity function discussed in Eq. (6)

It is to be noted here, that the community detection scheme used here is a one-shot unsupervised learning scheme. Figure 11(a) shows that the top three nodes (nodes 1 to 3) are in same community, while the bottom one (node 4) belongs to different community. This result is strikingly similar to the table depicted in Fig. 3(b), where the values of centrality measures show similar trends, as discussed in Sect. 3.1.1. For structure-II, it also produced convincing results that are qualitatively similar to the results obtained by node-centrality measures. For structures-III and IV (Fig. 11 (c-d)), the interior is homogenous, defect-free, and consists of identical lattice points. So this method did not split them into multiple false-partitions: therefore producing very attractive results despite being one-shot learning method. Since the boundaries are itself defect-regions, false-partitioning is not unexpected. Figure 11(e) correctly identifies all the point defects and the regions which are defect-free. Figure 11(f), produces strikingly good results, which identify the three main regions separated by the line defects. This result is remarkable since machine learning methods do not have access to position information of the particles (only to connectivity or adjacency matrix; it is because of making the visual correlation we plot the particle using the position rather than as abstract graphs). It is even more interesting because this method is very robust. This statement deserves further qualifications. The eigenvalue method is not robust in the sense that eigenvalue decomposition cannot be always found for any array matrix. The matrix $A$ can be singular for a variety of reasons. For example, if we mistakenly double-count a particular node (making a duplicate entry), the resultant matrix will be singular, and the eigenvalue decomposition method will fail to yield any result. Also, if the system size is large, typically, eigenvalue calculation will not be practically feasible, but the present method will still be workable with minor adjustments/tweaking. These are the main advantages of the present method.

## 4 The relevance of present analysis in the context of deep learning and beyond

Let us also consider the supervised learning regime. One of the most exciting and prolific areas in this regime is deep learning. Architecturally, deep learning is a composite structure of linear graph networks and non-linear activation functions [51-54]. It is to be noted that the nonlinear functions are chosen upfront (generally not selected by training methods), and only the weights of the linear graph are optimized through proper training methods. It is also important to note that, though the non-linear activation functions impart tremendous flexibility, so much so that it satisfies the universality theorem (i.e., it can mimic any complex function with arbitrary accuracy), the actual shapes of these non-linear function are not important in the limit of (potentially) infinite number of layers. Therefore these functions can be arbitrarily chosen upfront [55], though some functions might turn out to be more appropriate for a given problem than others, or even might be chosen based on convenience alone. For example, the "sigmoid" function is popular because its derivatives are extremely easy to calculate; or even the most simplistic function such as "ReLu" is mostly favored these days. Since the actual shapes of these functions are not critically important, we are at liberty to take any non-linear function. Here we chose simple "On-Off" switches (these are essentially "perceptons") to develop the intuition. In this formalism, a large deep-learning network can be described as a superposition of many small linear sub-graphs, where these sub-graphs are connected/disconnected as by the suitable settings of this On-Off switches kind of activation functions and therefore solves a complex problem by "divide and conquer" approach, discussed next.

One of the most important approaches to a very "large and complex" problem is to divide it into "numerous" simple and manageable problems and to find the solution for each individual piece, which is sometimes loosely referred as "divide and conquer" approach. There are many such theoretical frameworks, such as the Fourier transform, the spherical harmonics, etc. [56, 57]. Generally, they differ from each other in the formulation of the protocols in the divide and conquer (or deconstruction and reconstruction) of the problem and solving it. Since a deep learning network resembles the superposition of many linear graphs connected/disconnected by the activation functions, the first step for understanding that network is to understand the basic properties/functionalities of the underlying linear subgraph. The present study illustrates how to achieve this not only through a mathematical operation but simultaneously developing a visual analog of this. The bigger challenge is the development of the entire mathematical/visual formalism for comprehensive analysis.

Deep learning networks employ sequential linear and non-linear operations. The divide between these operations can be subtle in various instances. Indeed, the evolution of all physical systems (whether these are, on a large scale, linear or non-linear) is, at the basic level, quantum mechanical. Since quantum mechanics is a linear theory (with the evolution being a unitary linear map) adhering to the (linear) Schrodinger equation, all non-linearities in observed physical systems ultimately arise from effective truncation of one sort or another of a system that evolves unitarily in time. The individual layers in a deep learning network may be viewed as discrete time slices of a progressively truncated physical system that evolves in time. In the quantum mechanical setting, truncation by "integrating out" (summing over part of the system so that it becomes "hidden") gives rise to non-linear temporal evolution in the remaining degrees of freedom that are not summed over. Such an "integration" yields non-linearities and possible effective dissipation in various many body systems. Indeed, microscopically, energy is conserved. However, if one only analyzes part of the system then the energy (and other quantities) need not be conserved in a non-unitary evolution. Even the evolution of a single quantum mechanical particle described by the Schrodinger equation can display expectation values that solve non-linear (typically essentially classical) equations of motion. When computing an average, one examines only part of the full space of particle states (i.e., one number instead of its full Hilbert space) and in line with the qualitative discussion above the resulting evolution need not be linear. In the quantum arena (and trivially elsewhere), general non-linear functions of matrices naturally arise when describing the system evolution in time. These further trivially give rise to non-linearities as the system progress from one time slice to the next. The current article focused on classical particle systems. In a future publication, we will explore the effective non-linearities arising in quantum system in the context of deep networks

## 5 CONCLUSIONS

ML techniques are rapidly changing both the scientific and commercial world at an unthinkable pace. However, a deeper understanding of this is critically lacking, which might hamper its future progress. It is therefore imperative to develop scientific intuition behind the ML processes. Towards this end, this paper used six different 2-D materials systems as test cases, though the finding can be easily generalized to other domains of computation also. In essence, this article demonstrates few key ideas and their generic applications in ML through the use of those six test cases. First, we demonstrate the concept of node-centrality, using the principal eigenvector as a tool for classification, establishing connection between the solutions of wave equations and eigenvectors. The consequence of defects on the changes in the pattern of principal eigenvectors are depicted. Though this is not a particularly attractive method for ML because all matrices might not produce eigenvalues and eigenvectors, it produces scientifically rigorous results whenever applicable. This is of immense importance for the development of intuition which is beneficial for understanding other more complex methods. Second, we investigate spectral decomposition for these six structures. Third, the clustering method are described which produces very exciting results. The intuition for this method is developed with the help of the concepts developed by the previous methods. In the end, a fine underlying mathematical connection between the unsupervised methods and supervised methods (like deep learning) are discussed.

# Supplementary Information

**S1: Discrete Element Method (DEM) Simulation protocol**

**S1.1 DEM Algorithm**

The algorithm of the DEM simulation is given below. The pictorial representation of the flow chart of the DEM algorithm used in the present study is given in Fig. S1.1. Following steps are performed:

1. *Initialization*: In this step, the important parameters which are required for simulation are initialized. These parameters include coordinates of the particles, and their time derivatives, forces acting on each particle at the start of the simulation. The other particle parameters like its size, mass, type, coefficient of friction, Young modulus, etc. are also initialized.

2. *Predictor:* This step predicts the coordinates and time derivatives of the particles for the next time step using their previous time step values. For this purpose, Taylor series expansion is used which is given in eq. (s1.1a-c).

$$\vec{r}_i^{pr}(t+\Delta t) = \vec{r}_i(t) + \Delta t V_i(t) + \frac{1}{2}\Delta t^2 \frac{d^2 r}{dt^2} + \frac{1}{6}\Delta t^3 \frac{d^3 r}{dt^3} + \ldots \quad \text{(s1.1a)}$$

$$\vec{V}_i^{pr}(t+\Delta t) = V_i(t) + \Delta t \frac{d^2 r}{dt^2} + \frac{1}{2}\Delta t^2 \frac{d^3 r}{dt^3} + \ldots \quad \text{(s1.1b)}$$

$$\vec{r}_i^{pr}(t+\Delta t) = \frac{d^2 r}{dt^2} + \Delta t \frac{d^3 r}{dt^3} + \ldots \quad \text{(s1.1c)}$$

3. *Verlet list generation:* The basic property of mesoscopic particle dynamics is that neighbourhood relations between particles remain same for at least few integration time steps. So the information of neighbourhood of each particle will reduce the overall computation time[1]. Verlet list contains the information of neighbours of each particle. The neighbour of any particles is decided based on the distance between their surfaces. If this distance is smaller than a predefined constant, we assume that particle as its neighbour. Mathematically it can be represented as

$$\left|\vec{r}_i - \vec{r}_j\right| - \left(R_i + R_j\right) \leq \text{predefined distance} \tag{s1.2}$$

4. *Force calculation:* After predicting the next time step values of position and its time derivative in predictor step, force between interacting particles are calculated. It has two steps:

    (i) *Selection of the interaction pairs*: The interacting pairs are selected based on the verlet list generated in step (3). Since in granular materials, the long range forces are absent, the particles interact only in their neighbourhood.

    (ii) *Normal and tangential force calculation*: Pairwise interaction forces are calculated using different contact models. In the present study, modified Hertz model[2] for viscoelastic spheres[3] is used for calculation of normal force component of elastic spheres. The tangential component is calculates using Haff and Werner model[4].

The total force acting on a particle is the resultant of the normal and tangential forces of all the interacting particles and any externally applied force.

$$F^{total} = \sum_{j=1}^{k} \left( \vec{F}_{ij}^n \vec{e}_{ij}^n + \vec{F}_{ij}^t \vec{e}_{ij}^t \right) + F^{ext} \tag{s1.3}$$

Where, $F^n$ is the normal force acting on the particle and $k$ is the number of neighbors of the particle. $F^t$ is the tangential force acting on the particle and $F^{ext}$ is the externally applied force. $\vec{e}_{ij}^t$ is the unit vector perpendicular to the line joining the centre of the $i^{th}$ and $j^{th}$ particles and $\vec{e}_{ij}^n$ is the unit vector in the direction of line joining the centre of the $i^{th}$ and $j^{th}$ particles. Using this total force value acting on the particle, one can easily calculate the acceleration value as

$$\vec{r}^{corr} = \frac{F^{total}}{m} \tag{s1.4}$$

Where, $\vec{r}^{corr}$ is the correct value of acceleration of the particle, $m$ is the mass of the particle.

5. *Corrector:* Once the total force acting on each particle is known, we can calculate the acceleration acting on it. This acceleration value $\vec{r}^{corr}$ can be used to calculate the deviation in the predicted acceleration value in step (4) given in eq. (s1.5). This deviation is further used to correct the predicted position and its time derivative using the eq. (s1.6).

$$\Delta \vec{\ddot{r}} = \vec{\ddot{r}}_i^{corr} - \vec{\ddot{r}}_i^{pr} \tag{s1.5}$$

Where, $\vec{\ddot{r}}^{corr}$ is the correct acceleration obtained using eq. (s1.4) and $\vec{\ddot{r}}^{pr}$ is the predicted value of acceleration using Taylor series expansion in eq. (s1.1c)

$$\begin{pmatrix} \vec{r}_i^{corr}(t+\Delta t) \\ \vec{V}_i^{corr}(t+\Delta t) \\ \vec{\ddot{r}}_i^{corr}(t+\Delta t) \\ \vec{\dddot{r}}_i^{corr}(t+\Delta t) \\ \bullet \\ \bullet \\ \bullet \\ \bullet \end{pmatrix} = \begin{pmatrix} \vec{r}_i^{pr}(t+\Delta t) \\ \vec{V}_i^{pr}(t+\Delta t) \\ \vec{\ddot{r}}_i^{pr}(t+\Delta t) \\ \vec{\dddot{r}}_i^{pr}(t+\Delta t) \\ \bullet \\ \bullet \\ \bullet \\ \bullet \end{pmatrix} + \begin{pmatrix} C_0 \\ C_1 \dfrac{1}{\Delta t} \\ C_2 \dfrac{2}{\Delta t^2} \\ C_3 \dfrac{6}{\Delta t^3} \\ \bullet \\ \bullet \\ \bullet \\ \bullet \end{pmatrix} \bullet \dfrac{\Delta t^2}{2} \Delta \vec{\ddot{r}} \tag{s1.6}$$

The coefficient $c_i$ depends on the order of algorithm used and also on the type differential equation used. In the current simulation algorithm of fifth order is used for which the values of $c_i$ are as given below-

$C_0 = 19/90$, $C_1 = 3/4$, $C_2 = 1$, $C_3 = 1/2$, $C_4 = 1/12$ (s1.7)

6. *Data extraction*: The desired data are recorded and can be extracted in files in regular time steps for further statistical analysis.

7. *Program termination*: The program can be terminated upon occurrence of a certain event. This event can be a predefined simulation time, total kinetic energy of the system falling below some threshold value indicating no further appreciable kinetic evolution of the system or packing density reaching certain pre-set value of interest etc. Otherwise the simulation is continued at step 2 and so on.

Simulation is terminated after a predefined time steps in all cases. All the important particle parameters used for this simulation are listed in table S1.1

**Table S1.1: Particle Properties.**

| | | |
|---|---|---|
| Young's modulus | $Y$: | $10^9$ Pa |
| Coefficient of friction | $\mu$: | 0.50 |
| Damping constant | $A$: | 0.01 sec |
| Tangential damping constant | $\gamma^t$: | 10 Nsec/m |
| Material density | $\rho_m$: | 8 g/cm$^3$ |
| Integration time step | $\Delta t$: | $10^{-6}$ sec |

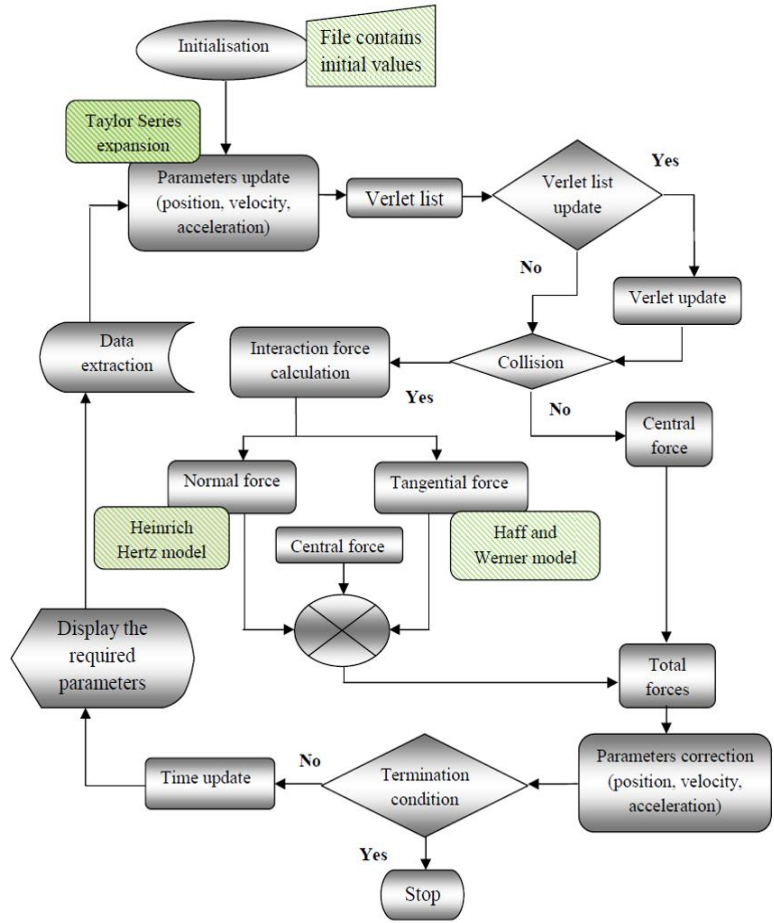

Fig. s1.1: Flow chart of the DEM algorithm used in present study.

**Simulation protocol:**

In the present simulation, particles distributed in a 2-D box with low initial density (~12%packing density) are subjected to an externally applied centripetal force (magnitude set equal to gravitational force) directed towards the centre of the box. Due to this applied force, particles move towards the centre of the box. The particles lose energy during collisions among themselves owing to the friction and damping, which are typical for realistic granular particles. Finally, a dense packing is obtained. There are several research papers indicated the influence of the wall on final packed structure. Due to the wall, particles near it are arranged in ordered structure and this ordering goes nearly a depth of 5d to 6d from the wall towards centre of the box[5-6]. This type of packing is formed without being influenced by the container walls. Therefore they play a central role in understanding the fundamentals of geometric aspects of the packing itself. They are of huge importance to model many real life phenomena, like the homogenous nucleation, etc. More details can be found in [7].

## S2: Even Eigenvectors

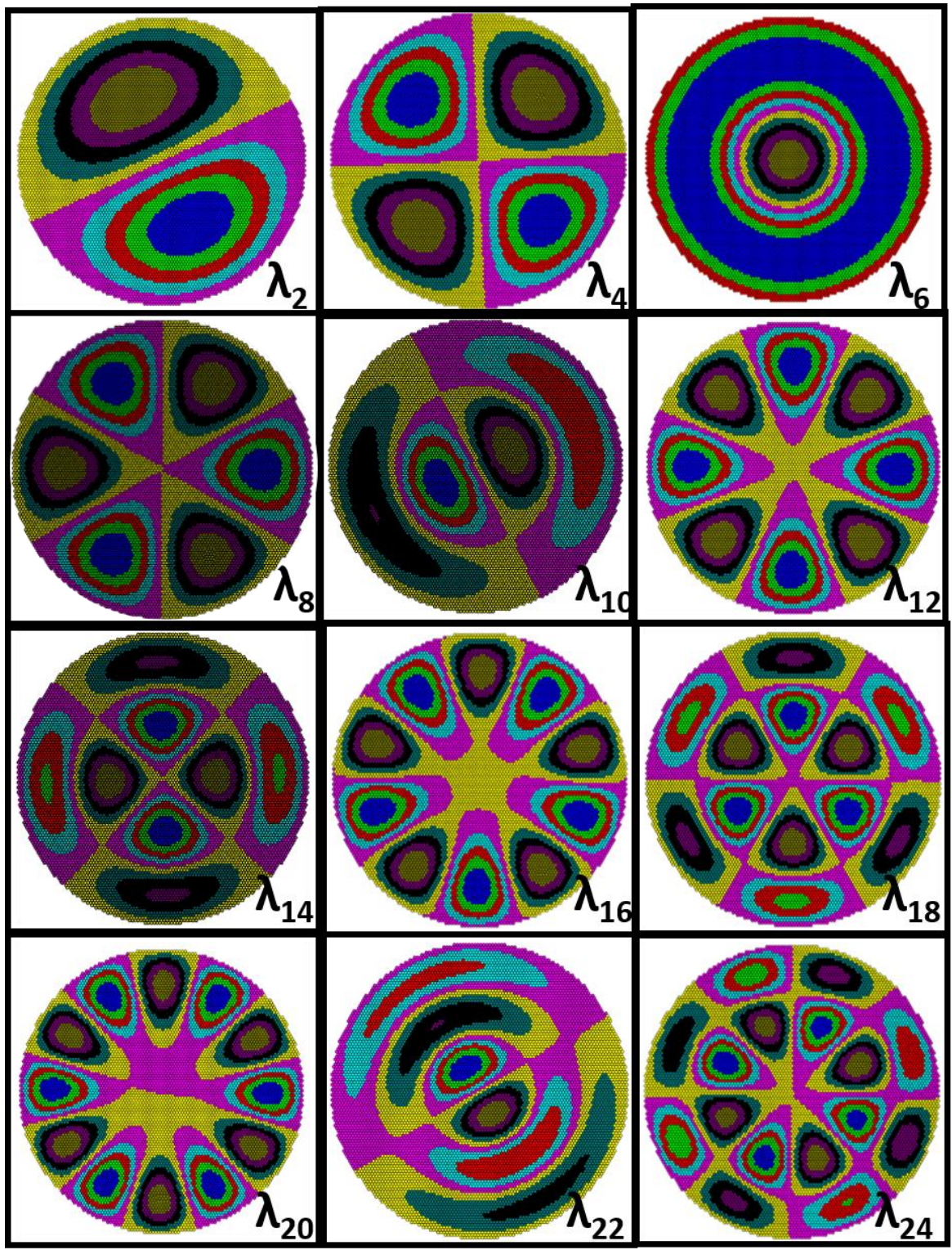

Fig. S2.1: The visual representation of eigenvectors corresponding to even eigenvalues (eigenvectors corresponding to the odd eigenvalues are shown in main article).

# S3: Reference values for color coding of nodes

**Table S3.1** Reference values for color-coding of nodes in each panel of figure 6 of the main article.

| Color-bar | V($\lambda_1$) | V($\lambda_3$) | V($\lambda_5$) | V($\lambda_7$) | V($\lambda_9$) | V($\lambda_{11}$) | V($\lambda_{13}$) | V($\lambda_{15}$) | V($\lambda_{17}$) | V($\lambda_{19}$) | V($\lambda_{21}$) | V($\lambda_{23}$) |
|---|---|---|---|---|---|---|---|---|---|---|---|---|
| olive | 0.0189 | 0.0190 | 0.0190 | 0.0190 | 0.0255 | 0.0197 | 0.0236 | 0.0131 | 0.0206 | 0.0230 | 0.0207 | 0.0306 |
| purple | 0.0170 | 0.0148 | 0.0148 | 0.0148 | 0.0198 | 0.0153 | 0.0184 | 0.0771 | 0.0161 | 0.0179 | 0.0161 | 0.0238 |
| black | 0.0150 | 0.0105 | 0.0105 | 0.0105 | 0.0142 | 0.0109 | 0.0131 | 0.0023 | 0.0117 | 0.0128 | 0.0115 | 0.0170 |
| teal | 0.0130 | 0.0063 | 0.0063 | 0.0063 | 0.0085 | 0.0066 | 0.0079 | -0.0030 | 0.0072 | 0.0077 | 0.0069 | 0.0102 |
| yellow | 0.0111 | 0.0021 | 0.0021 | 0.0021 | 0.0028 | 0.0022 | 0.0026 | -0.0084 | 0.0028 | 0.0026 | 0.0023 | 0.0034 |
| magenta | 0.0091 | 0.0021 | -0.0021 | -0.0021 | 0.0028 | -0.0022 | -0.0026 | -0.0138 | -0.0017 | -0.0026 | -0.0023 | -0.0034 |
| cyan | 0.0071 | 0.0063 | -0.0063 | -0.0063 | 0.0085 | -0.0066 | -0.0079 | -0.0192 | -0.0061 | -0.0077 | -0.0069 | -0.0102 |
| red | 0.0052 | 0.0105 | -0.0105 | -0.0105 | 0.0142 | -0.0109 | -0.0131 | -0.0245 | -0.0106 | -0.0128 | -0.0115 | -0.0170 |
| green | 0.0032 | 0.0148 | -0.0148 | -0.0148 | 0.0198 | -0.0153 | -0.0184 | -0.0299 | -0.0151 | -0.0179 | -0.0161 | -0.0238 |
| blue | 0.0012 | 0.0190 | -0.0190 | -0.0190 | -.0255 | -0.0197 | -0.0236 | -0.0353 | -0.0195 | -0.0230 | -0.0207 | -0.0306 |